\documentclass[sigconf,natbib=true,anonymous=false]{acmart}
\usepackage{multirow}
\usepackage{makecell}
\usepackage{adjustbox}
\AtBeginDocument{%
  \providecommand\BibTeX{{%
    \normalfont B\kern-0.5em{\scshape i\kern-0.25em b}\kern-0.8em\TeX}}}

\setcopyright{acmlicensed}
\copyrightyear{2022}
\acmYear{2022}
\acmDOI{10.1145/3477495.3532078}

\acmConference[SIGIR '22]{Proceedings of the 45th International ACM SIGIR Conference on Research and Development in Information Retrieval}{July 11--15, 2022}{Madrid, Spain.}
\acmBooktitle{Proceedings of the 45th International ACM SIGIR Conference on Research and Development in Information Retrieval (SIGIR '22), July 11--15, 2022, Madrid, Spain}

\acmPrice{15.00}
\acmISBN{978-1-4503-8732-3/22/07}

\begin{document}

\title{Video Moment Retrieval from Text Queries via Single~Frame~Annotation}

\author{Ran Cui}
\authornote{Both authors contributed equally to this research.}
\affiliation{%
  \institution{The Australian National University, Australia}
  \country{}
}
\email{ran.cui@anu.edu.au}

\author{Tianwen Qian}
\authornotemark[1]
\affiliation{%
  \institution{Fudan University, China}
  \country{}
}
\email{twqian19@fudan.edu.cn}

\author{Pai Peng}
\authornote{Corresponding authors.}
\affiliation{%
  \institution{bilibili, China}
  \country{}
}
\email{pengpai@bilibili.com}

\author{Elena Daskalaki}
\affiliation{%
  \institution{The Australian National University, Australia}
  \country{}
}
\email{eleni.daskalaki@anu.edu.au}

\author{Jingjing Chen}
\authornotemark[2]
\affiliation{%
  \institution{Fudan University, China}
  \country{}
}
\email{chenjingjing@fudan.edu.cn}

\author{Xiaowei Guo}
\affiliation{%
  \institution{bilibili, China}
  \country{}
}
\email{weide@bilibili.com}

\author{Huyang Sun}
\affiliation{%
  \institution{bilibili, China}
  \country{}
}
\email{sunhuyang@bilibili.com}

\author{Yu-Gang Jiang}
\affiliation{%
  \institution{Fudan University, China}
  \country{}
}
\email{ygj@fudan.edu.cn}

\renewcommand{\shortauthors}{Ran Cui and Tianwen Qian, et al.}

\begin{abstract}
 Video moment retrieval aims at finding the start and end timestamps of a moment (part of a video) described by a given natural language query. \textit{Fully supervised} methods need complete temporal boundary annotations to achieve promising results, which is costly since the annotator needs to watch the whole moment. \textit{Weakly supervised} methods only rely on the paired video and query, but the performance is relatively poor. In this paper, we look closer into the annotation process and propose a new paradigm called ``glance annotation''. This paradigm requires the timestamp of only one single random frame, which we refer to as a ``glance'', within the temporal boundary of the fully supervised counterpart. We argue this is beneficial because comparing to weak supervision, trivial cost is added yet more potential in performance is provided. Under the glance annotation setting, we propose a method named as \textit{\textbf{Vi}deo moment retrieval via \textbf{G}lance \textbf{A}nnotation} (ViGA)\footnote{We release our codes and glance re-annotated datasets at \textit{https://github.com/r-cui/ViGA}.} based on contrastive learning. ViGA cuts the input video into clips and contrasts between clips and queries, in which glance guided Gaussian distributed weights are assigned to all clips. Our extensive experiments indicate that ViGA achieves better results than the state-of-the-art weakly supervised methods by a large margin, even comparable to fully supervised methods in some cases.
\end{abstract}



\maketitle

\fancyhead{} 
\section{Introduction}
\begin{figure}
    \centering
    \includegraphics[width=\linewidth]{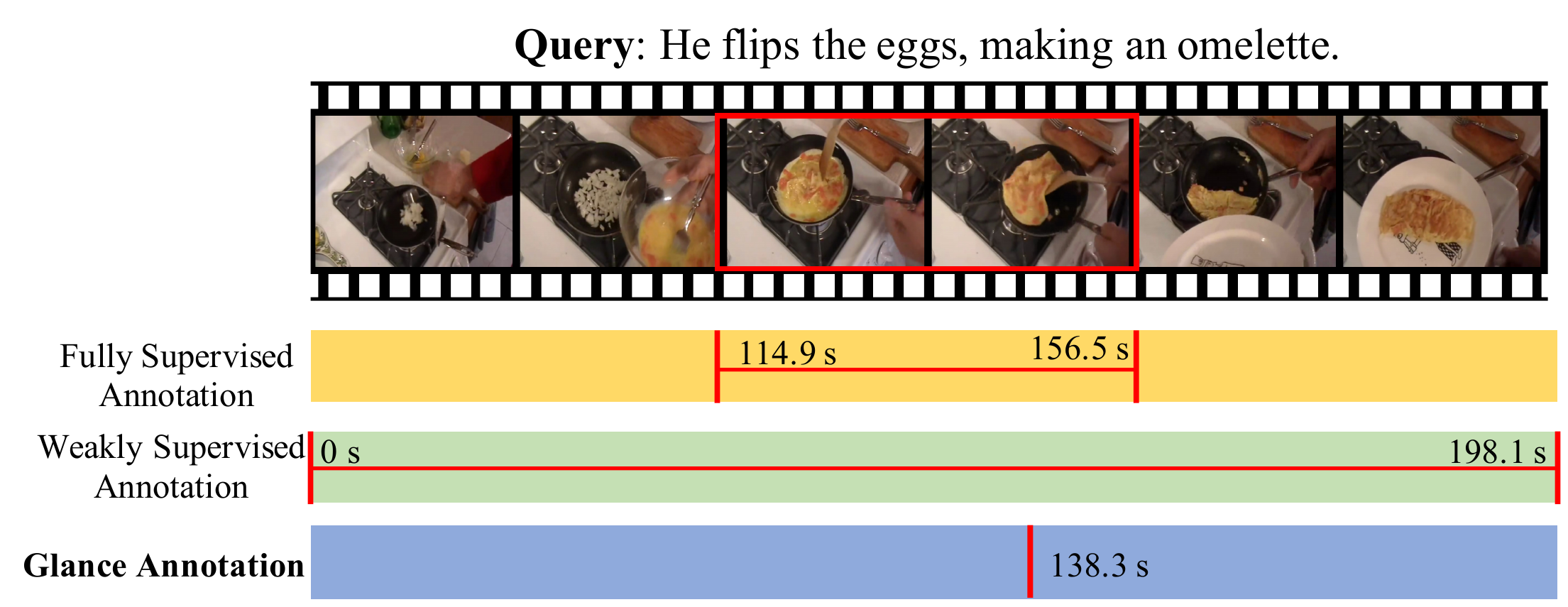}
    \caption{Illustration of different VMR training example annotation paradigms. The fully supervised setting marks the start and end timestamps of the moment corresponding to query. Weak supervision only annotates the video-text pair. Our proposed glance annotation marks a single timestamp in the moment.}
    \label{fig:intro_fig}
\end{figure}

Video moment retrieval (VMR), initially proposed in \cite{gao2017tall, anne2017localizing}, is the task of retrieving the segment described by a given natural language query from an untrimmed video. This task, also known as natural language video localization \cite{zhao2017temporal, liu2021survey, zhang2020span} and video temporal grounding \cite{chen2018temporally, mun2020local}, is a fundamental problem in computer vision understanding and visual information retrieval. Differing to an earlier task video action localization \cite{escorcia2016daps, lin2018bsn}, which aims at localizing pre-defined categorized actions from the video, VMR is considered as a more difficult task since the query is generalized to free natural language thus involving with more complex cross-modal contents understanding. VMR can be widely applied in many scenarios such as video browsing websites and semantics based video search engines.

To date, deep learning methods have approached VMR from two directions differing in the way of data annotation. In building a dataset of \textit{fully supervised VMR}, given the target video, the annotator is asked to choose a segment in the video and write a short text query to describe the segment. In the meanwhile, the start and end timestamps of this segment are noted down. Thus, one example in the dataset is a quadruplet of video, query, start and end, denoted by $(V,Q,st,ed)$. Though many methods under \textit{fully supervised VMR} \cite{gao2017tall, xu2019multilevel, zhang2020learning, mun2020local, zhang2020span, chen2021end, wang2021structured} have achieved good performance, an obvious disadvantage of this data annotating paradigm is its high time cost. Besides, the annotation quality varies according to the annotator's subjective judgements, especially in determining the start and end: the annotator is forced to give specific timestamps of the query, but the video segment is often not separated to its context with clear border. For example, to annotate the start and end of a query ``the man turns the light on'', one might consider the very second that the switch is toggled should be the temporal interval, but another might start the annotation from when the man walks towards the switch. This makes \textit{fully supervised VMR} prone to subjective annotation noise. To avoid these disadvantages, \textit{weakly supervised VMR} is proposed \cite{mithun2019weakly}, in which one example in the dataset is simply $(V,Q)$, and no start nor end annotation is available. Though not comparable to \textit{fully supervised VMR} in performance, many studies \cite{lin2020weakly, wu2020reinforcement, ma2020vlanet, song2020weakly, tan2021logan, huang2021cross} have shown that weak supervision is a feasible compromise when the annotating resources are limited. 

In our study, we argue that weak annotation can be augmented with trivial cost and propose ``glance annotation'', a new paradigm of data annotation in VMR. As illustrated in Figure \ref{fig:intro_fig}, a training example under glance annotation is composed of $(V,Q,g)$ in which $g$ is any timestamp between $st$ and $ed$. This paradigm is motivated by the fact that to annotate even a weak example, it is still inevitable for one to watch the video in order to write the query, and very often it is sufficient to know what the moment is about by watching only a short snippet of it. Assuming that with properly designed graphical user interface to support the annotation, one can note down an instant timestamp $g$ during ``glancing'' the video with no more effort than a mouse click. Glance annotation not only largely saves the time consumption in watching the video compared to full supervision, but also provides more information than weak supervision.

To validate the usability of glance annotation, we re-annotate three publicly available fully supervised VMR datasets, namely ActivityNet Captions \cite{krishna2017dense}, Charades-STA \cite{gao2017tall} and TACoS \cite{tacos2013} by substituting $st$ and $ed$ with a \textit{uniformly} sampled timestamp $g$ in range $[st, ed]$. Under this setting, we propose a contrastive learning based method named \textit{Video moment retrieval via Glance Annotation} (ViGA). Due to the lack of $st$ and $ed$, ViGA follows the multiple-instance learning (MIL) strategy widely adopted in \textit{weakly supervised VMR}, which uses the correspondence between $V$ and $Q$ as the supervision signal. In doing the \textbf{training}, the main objective is to obtain a proper cross-modal encoder to project $V$ and $Q$ to a joint embedding space, which satisfies that the distance between the embeddings of corresponding $(V,Q)_\text{pos}$ is closer and the distance between the embeddings of other combinations $(V,Q)_\text{neg}$ is farther. ViGA extends this idea by splitting $V$ into multiple clips $C$ and learning in the granularity $(C,Q)$ instead, for making use of $g$ by enabling an assignment of different weights to all clips. Specifically, we use heuristic Gaussian distribution peaking at the glance position to generate the weights. In doing the \textbf{inference}, we follow the common proposal-based inference as in many weakly supervised methods, yet adjust the classical sliding window proposal generation to an anchor driven proposal generation to better fit our training strategy. To be specific, those sliding window proposals not including a first selected anchor frame are pruned out. We enable the network finding the anchor by adding an additional training objective of focusing the attention of our multihead-attention \cite{vaswani2017attention} based backbone to the glance position.

As will be shown in the rest of this paper, ViGA significantly outperforms the state of the art of \textit{weakly supervised VMR}, even comparable to \textit{fully supervised VMR} when a high precision of retrieved segment is not required. Our contributions are summarized as follows:

\begin{itemize}
\item We propose glance annotation, a new VMR annotating paradigm, which requires no more significant annotating effort than weakly supervised VMR, while provides more potential regarding the retrieval performance.
\item We propose a contrastive learning based method ViGA for glance annotated VMR, which achieves significantly better performance than weakly supervised methods.
\item We release a unified version of glance annotations on top of three publicly available datasets ActivityNet Captions \cite{krishna2017dense}, Charades-STA \cite{gao2017tall} and TACoS \cite{tacos2013}, to encourage future research on this topic.
\end{itemize}

\begin{figure*}
    \centering
    \includegraphics[width=0.9\textwidth]{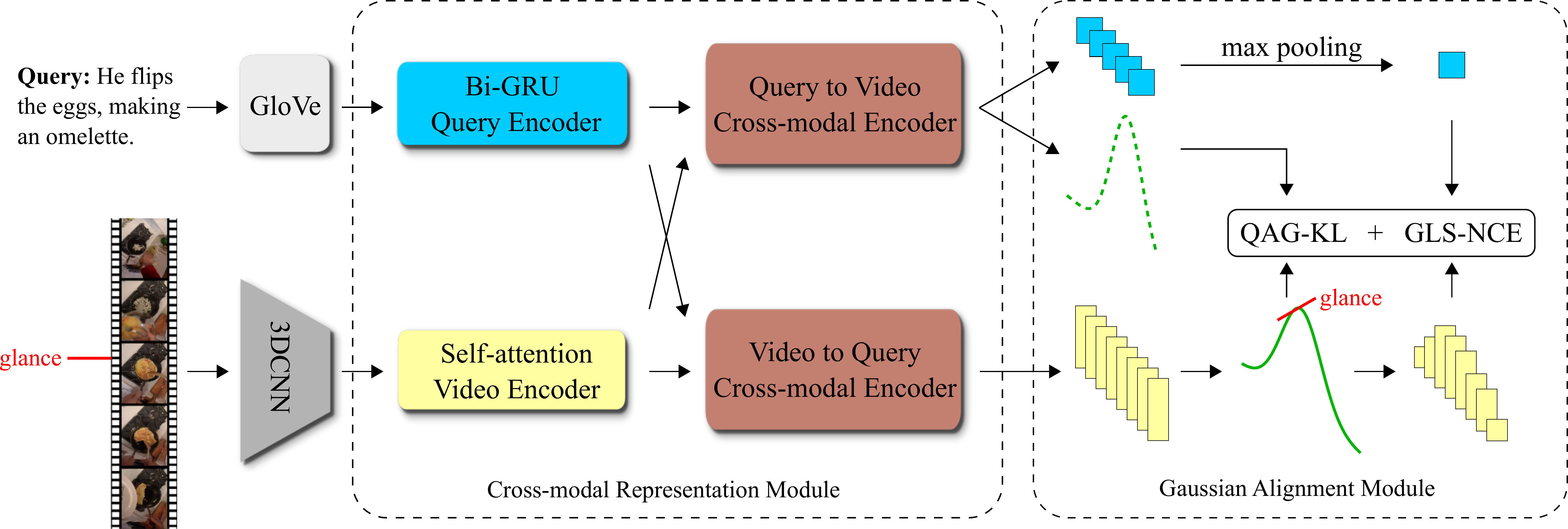}
    \caption{Illustration of our training framework. In the \textit{Gaussian Alignment Module}, a blue square denotes a word feature and a yellow rectangle denotes the feature of one video frame. We use different heights of yellow squares to illustrate the different weights of the frames. The solid and dashed green curves represent the heuristic Gaussian distribution generated with the glance and the attention distribution generated by the model, respectively.}
    \label{fig:framework}
\end{figure*}

\section{Related Work}
\label{sec:2}
After initially proposed by \cite{gao2017tall, anne2017localizing}, early VMR studies mostly use the annotated start and end timestamps for the video-text temporal alignment learning, which we term as \textit{fully supervised VMR} \cite{gao2017tall, anne2017localizing, liu2018cross, jiang2019cross, chen2018temporally, zhang2020learning, chen2019localizing, mun2020local}. Due to the expensive annotation cost, researchers then began to exploit on learning under weak annotation with video-text pairs only, which we term as \textit{weakly supervised VMR} \cite{duan2018weakly, mithun2019weakly, gao2019wslln, lin2020weakly, ma2020vlanet, zhang2020counterfactual, huang2021cross}.

\subsection{Fully Supervised VMR}
\label{sec2.1}
Existing \textit{fully supervised VMR} methods can be categorized into two groups. \textbf{Two-stage} methods \cite{gao2017tall, anne2017localizing, jiao2018three, liu2018cross, jiang2019cross, ge2019mac, xu2019multilevel} typically generate some pre-segmentation of proposal candidates using a sliding window or other proposal networks, then input the generated proposals and the text query separately into a cross-modal matching network to predict matching confidence and select the best matching segment as the output. Hendricks \textit{et al.} \cite{anne2017localizing} first proposed Moment Context Network (MCN), which generated proposals based on sliding window, and then projected the video moment feature and text query feature into a common representation space. Then they used $L2$ distance as a measure to optimize triplet loss to narrow the distance of positive samples and enlarge the distance of intra-video and inter-video negative samples. Xu \textit{et al.} \cite{xu2019multilevel} proposed Query-guided Segment Proposal Network (QSPN) for alleviating the huge computation burden caused by sliding window. Specifically, QSPN integrated query features into video features to obtain attention weights in time indexing, then combined it with 3D Region of Interest (ROI) pooling to obtain the sparse proposals. \textbf{End-to-end} models \cite{chen2018temporally, zhang2019man, zhang2020learning, wang2020temporally, yuan2019find, chen2019localizing, mun2020local} can be divided into anchor-based \cite{chen2018temporally, qu2020fine, wang2020temporally, zhang2020learning} methods and anchor free \cite{chen2019localizing, chen2020learning, mun2020local, yuan2019find, zhang2020span} methods, in which they differ in using / not using proposals in prediction, respectively. As a typical work in the anchor-based category, Zhang \textit{et al.} \cite{zhang2020learning} proposed 2D Temporal Adjacent Networks (2D-TAN) that modeled the relations between segments of varying durations using a two-dimensional feature map. The $(i, j)$-th location of the feature map indicated the start and end timestamps of the proposed segments. It then employed a Convolutional Neural Network (CNN) to model the contextual interaction between various segments, using ground truth labels to optimize the prediction score of each suggestion in the feature map. For anchor-free methods, they usually predict the probability of a frame being the start or end, or use a neural network to directly regress the values of start and end. For example, Lei \textit{et al.} proposed XML \cite{lei2020tvr} and used the 1D Convolutional Start-End detector (ConvSE) to generate the start and end scores on the late fused query-clip similarity matrix.

\subsection{Weakly Supervised VMR}
Although the fully supervised methods achieve good performance, the expensive cost of annotating the temporal boundary limits practical applications. Therefore, researchers recently began to pay attention to the \textit{weakly supervised VMR} \cite{duan2018weakly, mithun2019weakly, gao2019wslln, lin2020weakly, ma2020vlanet, zhang2020counterfactual, huang2021cross}. Under the weakly supervised setting, we cannot obtain the detailed start and end annotation of each query, only know whether the query and video is a positive pair during training stage. Under this constraint, most methods adopt the MIL framework. In MIL-based VMR, the model learns the video-text alignment at video-level by maximizing similarity scores of positive examples and suppressing them on negative examples. Text-Guided Attention (TGA) \cite{mithun2019weakly} was a typical pioneer work under the weak setting, which learned text-aware video representation and leverages ranking loss to distinguish positive and negative samples. Ma \textit{et al.} proposed VLANet \cite{ma2020vlanet} which attempted to eliminate some irrelevant suggestions in the process of MIL. Cross-sentence Relations Mining (CRM) \cite{huang2021cross} presented by Huang \textit{et al.} explored the temporal information modeling in MIL using combinational associations among sentences. Semantic Completion Network (SCN) \cite{lin2020weakly} provided another reconstruction-based idea of restoring the masked keywords in query according to visual proposal and context information for the alignment learning between modalities. Although weakly supervised VMR greatly reduces the burden of annotation, the performance of weak method has a significant gap between the fully supervised method on the test set. 

\section{Methodology}
\label{sec3:method}
 In this section, we first formally define the problem of glance annotated VMR and give an overview of our method ViGA. We then introduce the two modules which form our training pipeline in Section \ref{sec3.3} and \ref{sec:GAM}, respectively. The inference process is detailed in Section \ref{sec3.5}.

\subsection{Glance Annotation}
\label{sec3.1}
Given an untrimmed video $V$ and a text query $Q$ that semantically describes a segment of the video, the VMR task aims at finding the start and end timestamps $st$ and $ed$, such that moment $V_{st:ed}$ best corresponds to the query description. In \textit{fully supervised VMR}, complete human annotated $st$ and $ed$ information is provided. In contrast, under the \textit{weakly supervised VMR} setting, only aligned $(V, Q)$ pairs are available, with no fine-grained $st$ or $ed$ information. Our glance annotation scenario lies in between: a single timestamp $g$, satisfying $st\leq g\leq ed$, is available at the training stage. We refer to this timestamp $g$ as a ``glance''.

\subsection{Algorithm Overview}
\label{sec3.2}
Similar to the weakly supervised setting, it is not possible to let a network learn to directly output $st$ and $ed$ under glance annotation, due to the lack of complete supervision signals. Instead, our method selects a clip $C$ from $V$ that best matches $Q$ from a set of proposals as the output. To learn this visual-textual alignment, many studies in \textit{weakly supervised VMR} adopt the MIL strategy and turn into exploiting the correspondence of $(V, Q)$. Videos and queries that we know are from the same example are marked as positive correspondence $(V, Q)_\text{pos}$, while all other combinations in the batch are treated as negative $(V, Q)_\text{neg}$. Our work extends this idea to a finer-grained $(C, Q)$ level. Specifically, we build a network that projects inputs from textual and visual modalities to a joint embedding space, and train the network with a clip-to-query contrastive objective, which pulls the distance between $(C, Q)_\text{pos}$ closer and pushes the distance between $(C, Q)_\text{neg}$ farther.

\paragraph{\textbf{Training}}
The overall structure of our training pipeline is illustrated in Figure \ref{fig:framework}. After an initial feature extraction from pre-trained models, our \textit{Cross-modal Representation Module} encodes the two input modalities by first applying two independent uni-modal encoders, and then cross-interacting the two uni-modal features to each other. As a result, token-level (words for text and frames for video) cross-modal features are obtained. The \textit{Query to Video Cross-modal Encoder} additionally outputs an attention distribution across all video frames. To train the network, we propose a \textit{Gaussian Alignment Module}, in which we generate a heuristic Gaussian distribution peaking on the glance timestamp. All video frame features are weighted by this heuristic distribution in calculating our Gaussian Label-Smoothed Noise Contrastive Estimation loss (GLS-NCE). The same Gaussian heuristic distribution is further used in our Query Attention Guide Kullback–Leibler Divergence loss (QAG-KL) to guide the learning of our network. The total loss of our network is a fusion of the two loss functions.

\paragraph{\textbf{Inference}}
To align with the training design, we propose a corresponding Query Attention Guided Inference strategy. After the network forward pass up to the \textit{Cross-modal Representation Module}, the frame that gathers the most attention in \textit{Query to Video Cross-modal Encoder} is marked as the anchor frame. We sample proposals of different sizes around this anchor frame (\textit{i.e.}, a proposal must contain the anchor frame) and form a pool of proposals. The proposal that gets the highest dot-product similarity to the sentence feature is selected as the final output.

\subsection{Cross-modal Representation Module}
\label{sec3.3}
Given a video $V = [v_1, ..., v_{L_v}]$, and a query $Q = [q_1, ..., q_{L_q}]$, we encode deep features of the two inputs using the proposed \textit{Cross-modal Representation Module}. Specifically, we first use two independent encoders to ensure sufficient understanding of uni-modal semantics for video and query. Next, to enable the cross-modal learning, we fuse the semantics of the two modalities in the subsequent bidirectional cross-modal encoder. As a result, per-token representations $\mathbf{v} \in \mathbb{R}^{L_v \times d_\text{model}}$ and $\mathbf{q} \in \mathbb{R}^{L_q \times d_\text{model}}$ are obtained, where $d_\text{model}$ is the dimension of the joint embedding feature (and also the overall hidden dimension of our network).

\paragraph{\textbf{Query Encoding}} 
A bidirectional Gated Recurrent Unit (GRU) is applied to encode the sequential semantics of all $L_q$ words in $Q$, taking word embeddings from the pre-trained GloVe \cite{pennington2014glove} model as input. A word-level feature $\mathbf{q}_i$ is the concatenation of the forward and backward hidden states of the final layer of the GRU, given by

\begin{align}
    \mathbf{q}_i = [\overset{\rightarrow}{\mathbf{h}_i};\overset{\leftarrow}{\mathbf{h}_i}]\in \mathbb{R}^{d_\text{model}}.
\end{align}

\paragraph{\textbf{Video Encoding}} 
For an untrimmed video, we first extract features using a pre-trained CNN, such as C3D \cite{tran2014c3d}, I3D \cite{carreira2017quo} and VGG \cite{simonyan2014very}, followed by a fully connected layer to map the feature dimension to $d_\text{model}$. To encode the sequential semantics of the extracted video feature $\mathbf{v}$, we apply a multihead self-attention module \cite{vaswani2017attention} across all the frames. The encoding at the $i$-th frame is given by

\begin{align}
    \text{Attn}(Q(\mathbf{v}_i),K(\mathbf{v}),V(\mathbf{v})) = \text{softmax}(\frac{Q(\mathbf{v}_i)K(\mathbf{v})^T}{\sqrt{d_\text{model}/h}})V(\mathbf{v}),
\end{align}
where $Q(\cdot)$, $K(\cdot)$ and $V(\cdot)$ are three independent linear transformations from $d_\text{model}$ to $d_\text{model}$, and $h$ denotes the number of heads. 

\paragraph{\textbf{Cross-modal Encoding}}
To fuse the information from the two modalities, we apply cross-modal multihead attention after the individual uni-modal self encoding, \textit{i.e.}, using one modality as query and the other as key and value. In this way, the cross-encoding of the $i$-th word is given by

\begin{align}
    \text{Attn}(Q(\mathbf{q}_i),K(\mathbf{v}),V(\mathbf{v})) = \text{softmax}(\frac{Q(\mathbf{q}_i)K(\mathbf{v})^T}{\sqrt{d_\text{model}/h}})V(\mathbf{v}),
\label{eq:q2v}
\end{align}
and the cross-encoding of the $i$-th frame is given by

\begin{align}
    \text{Attn}(Q(\mathbf{v}_i),K(\mathbf{q}),V(\mathbf{q})) = \text{softmax}(\frac{Q(\mathbf{v}_i)K(\mathbf{q})^T}{\sqrt{d_\text{model}/h}})V(\mathbf{q}).
\label{eq:v2q}
\end{align}

For each encoding module in the uni-modal encoding and the cross-modal encoding, the module is subsequently followed by a two-layer feed-forward module activated by ReLU \cite{nair2010rectified} to further enhance the encoding capacity. Moreover, we follow the standard configuration of multihead attention modules, where layernorm \cite{ba2016layer}, dropout \cite{srivastava2014dropout}, position embedding \cite{devlin2018bert} and residual connection \cite{he2016deep} are applied.

\subsection{Gaussian Alignment Module}
\label{sec:GAM}
\begin{figure}
    \centering
    \includegraphics[width=\linewidth]{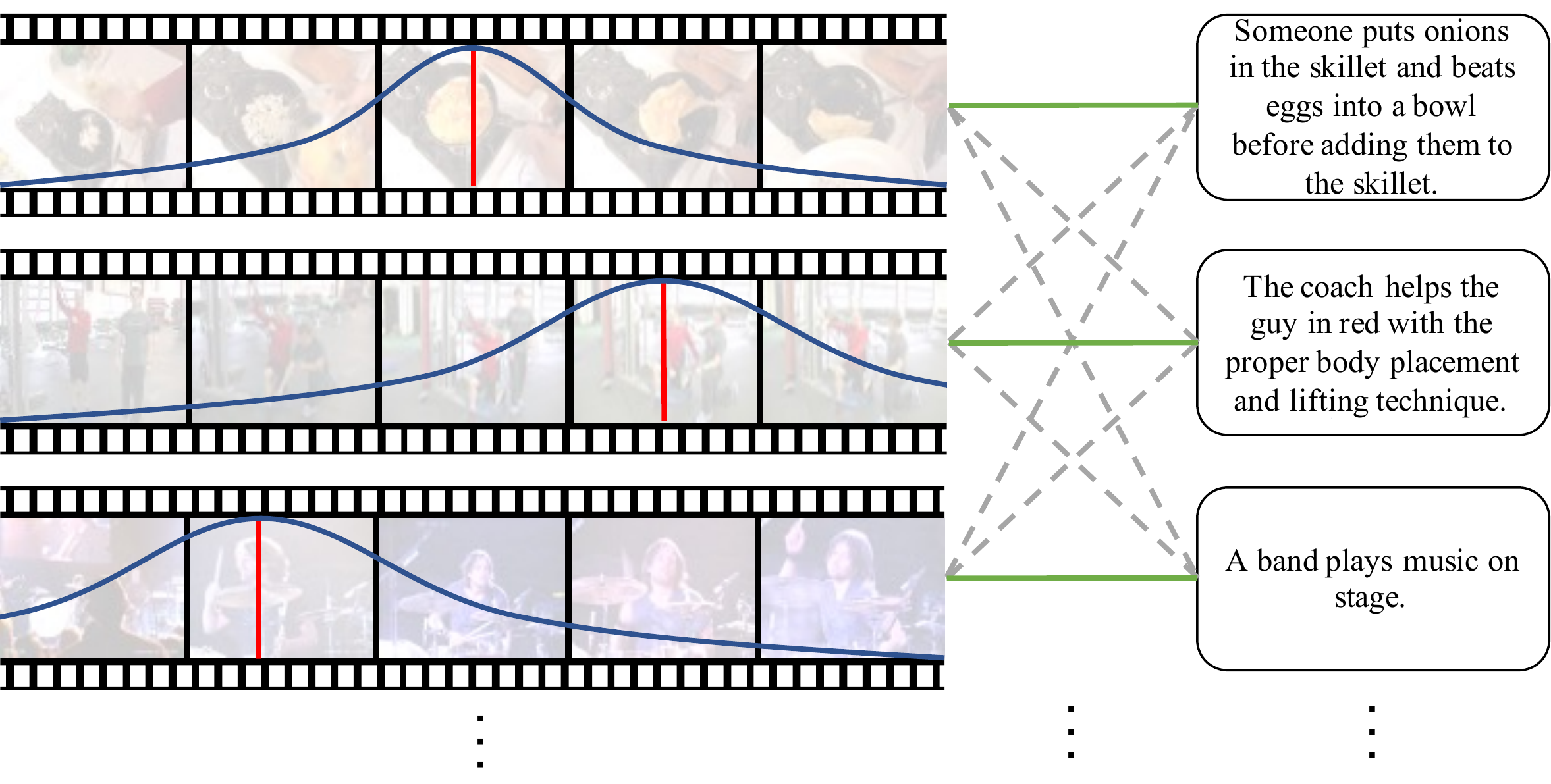}
    \caption{Illustration of clip-level MIL training strategy in one batch in the \textit{Gaussian Alignment Module}. Green solid lines indicate positive correspondences, and gray dashed lines indicate negative matching.}
    \label{fig:loss}
\end{figure}

In MIL-based methods under \textit{weakly supervised VMR}, the general paradigm is to learn proper deep representation $f_v(V) \in \mathbb{R}^{d_\text{model}}$ and $f_q(Q) \in \mathbb{R}^{d_\text{model}}$ that corresponding pairs align closer to each other via contrastive learning. We extend this idea of video-level MIL and propose a \textit{Gaussian Alignment Module} that transforms the problem to a finer-grained clip-level MIL to train the preceding \textit{Cross-modal Representation Module}. Our motivation is that the existence of glance $g$ makes frames in the video in-equally important in terms of the relevance to the query. For a frame $v_i$, the relevance is higher when its temporal distance to $g$ is closer: consider a long video including scene change, the frames that are too far away from $g$ might contain totally irrelevant semantics. Mathematically, Gaussian distribution has the characteristic that the highest probability value at the mean point and gradually reduces the probability to both sides, which aligns consistently to our motivation. Thus, we use Gaussian distribution to model this relevance. As illustrated in Figure \ref{fig:loss}, all video frames are assigned with Gaussian-distributed weights where the peak position of Gaussian is the glance $g$. To get the weight of the $i$-th frame, we scale the index $i\in \{1,2,...,L_v\}$ into the domain $[-1, 1]$ by linear transformation

\begin{align}
    f(i)=(i-1)\cdot \frac{2}{L_v-1}-1,
\end{align}
and sample the Gaussian values via the probability density function

\begin{align}
    G(i)=\text{norm}(\frac{1}{\sqrt{2\pi}\sigma}\text{exp}(-\frac{(f(i)-f(g))^2}{2\sigma^2})),
\label{eq:gaussian}
\end{align}
where $\sigma$ is a hyperparameter, and the normalization scales $G(i)$ where $i\in [-1, 1]$ into range $[0, 1]$.

After different weights are assigned across the video frames, we are able to get video clips with different weights as training examples. A sliding window of size $L_c$ with stride $s$ is applied on the video to get clips. Each clip is then max pooled along the frame dimension to generate the clip-level feature in the joint embedding space $\mathbb{R}^{d_\text{model}}$. To this end, the $i$-th clip feature $\mathbf{c}_i$ is given by

\begin{align}
    \mathbf{c}_i = \text{max\_pool}([\mathbf{v}_{(i-1) \cdot s + 1},..., \mathbf{v}_{(i-1) \cdot s+L_c}]) \in \mathbb{R}^{d_\text{model}}.
\label{eq:ci}
\end{align}
And each clip is assigned with a clip-level weight $w_{_i}$ sampled at the middle point of the clip, given by

\begin{align}
    w_{_i} = G((i-1) \cdot s + \frac{L_c}{2}).
\label{eq:wi}
\end{align}

Similarly, for the text modality, sentence-level feature $\mathbf{s}$ is obtained by max pooling its word-level features, given by

\begin{align}
    \mathbf{s} = \text{max\_pool}([\mathbf{q}_1,..., \mathbf{q}_{L_q}]) \in \mathbb{R}^{d_\text{model}}.
\end{align}

\paragraph{\textbf{GLS-NCE Loss}}
In \textit{weakly supervised VMR}, standard NCE loss on video level can be directly applied to train the video and query encoders $f_v(\cdot)$ and $f_q(\cdot)$ by contrasting $(V,Q)_\text{pos}$ against $(V,Q)_\text{neg}$ in one batch. For one video $V$ in a batch of $B$ video query pairs, there is only one positive matching query $Q^p$, and the rest $B-1$ queries are negative queries $Q^n$. Therefore, the standard video-level NCE loss is given by

\begin{align}
    \mathcal{L}_\text{Video-NCE}=-\text{log}(\frac{e^{f_v(V)^\top f_q(Q^p)}}{e^{f_v(V)^\top f_q(Q^p)} + \sum\limits^{B-1}_{i=1}e^{f_v(V)^\top f_q(Q^n_i)}}).
\label{eq:Video-NCE}
\end{align}
However, in our method, the proposed GLS-NCE loss is built on clip level. Each video is substituted by $N$ clips as in Equation \ref{eq:ci}. On this basis, for a video in the same batch of size $B$, clip-level NCE loss is extended to

\begin{align}
    \mathcal{L}_\text{Clip-NCE}=-\text{log}(\frac{\sum\limits^{N}_{i=1}e^{\mathbf{c}_i^\top f_q(Q^p)}}{\sum\limits^{N}_{i=1}e^{\mathbf{c}_i^\top f_q(Q^p)} + 
    \sum\limits^N_{i=1}\sum\limits^{B-1}_{j=1}e^{\mathbf{c}_i^\top f_q(Q^n_j)}
    }).
\label{eq:Clip-NCE}
\end{align}
Additionally, the clips also differ in weights given by Equation \ref{eq:wi}. To accommodate this, we implement clip-level NCE in the form of cross-entropy following MoCo \cite{he2020momentum}, and enable the weighting via label smoothing. In this GLS-NCE loss, the Gaussian weight $w_i$ of a clip $\mathbf{c}_i$ is assigned as the label smoothing amount, \textit{i.e.}, instead of using a one-hot label across the $B$ queries in the batch, we assign $w_i$ to the label of the positive query, and smooth the rest $B-1$ negative labels to $\frac{1-w_i}{B-1}$. In summary, for a clip $\mathbf{c}_i$ with weight $w_i$, its GLS-NCE loss is given by

\begin{align}
\mathcal{L}_\text{GLS-NCE}=w_i\cdot \text{log}(\mathbf{c}_i^\top Q^p)+    
  \sum\limits^{B-1}_{j=1}\frac{1-w_i}{B-1}\text{log}(\mathbf{c}_i^\top Q^n_j).
 \label{eq:GLS-NCE}
\end{align}

\begin{figure}
    \centering
    \includegraphics[width=\linewidth]{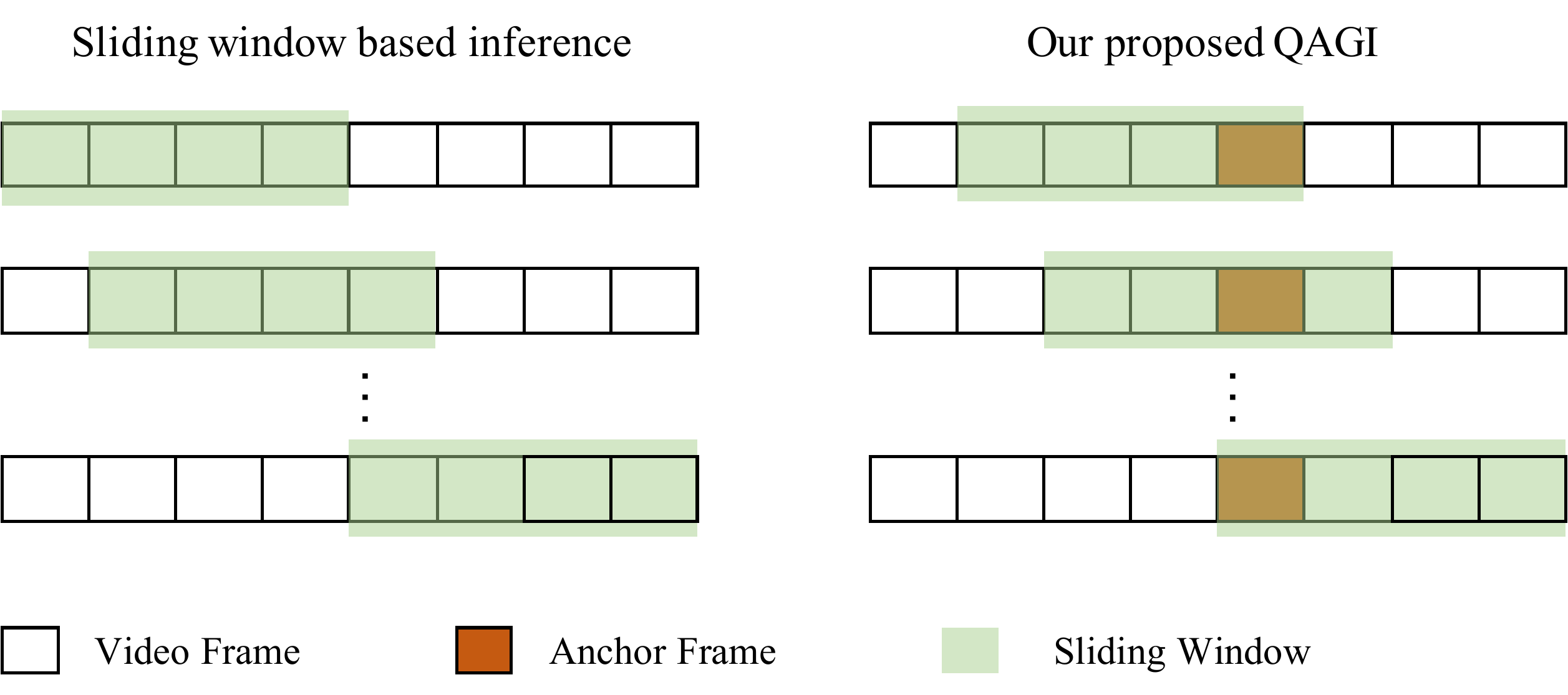}
    \caption{Comparison between sliding window based inference and our proposed \textit{Query Attention Guided Inference}.}
    \label{fig:eval}
\end{figure}

\paragraph{\textbf{QAG-KL Loss}}
To further smooth the learning and to align with the inference strategy to be explained in Section \ref{sec3.5}, we leverage the nature of attention mechanism \cite{vaswani2017attention} in our encoding module and propose the QAG-KL loss. Specifically, we use a KL divergence between the attention distribution of the Query to Video Cross-modal Encoder (Equation \ref{eq:q2v}) and the Gaussian guidance (Equation \ref{eq:gaussian}), to pull the attention distribution across all video frames closer to the Gaussian guidance. Since the query $Q$ contains $L_q$ words, we treat them equally and use the mean of their attention distributions as the sentence-level attention distribution. For the $i$-th frame in the video, the QAG-KL loss is given by

\begin{align}
    \mathcal{L}_\text{QAG-KL}&=G(i)(\text{log}G(i)-\mathbf{a}_i) \\
    \text{where}\;\; \mathbf{a}&=
    \frac{\sum\limits^{L_q}_{j=1}\text{softmax}(\frac{\mathbf{q}_j K(\mathbf{v})^T}{d_\text{model}/h})}{L_q} \in \mathbb{R}^{L_v}.
\label{eq:QAG-KL}
\end{align}

The complete loss function of a batch is the combination of the GLS-NCE loss across all clips in the batch and the QAG-KL loss across all frames of all videos in the batch, given by

\begin{align}
    \mathcal{L}=\sum\limits_{\mathbf{c}}\mathcal{L}_\text{GLS-NCE} + \sum\limits_{v}\sum\limits_{L_v}\mathcal{L}_\text{QAG-KL}.
\end{align}

\subsection{Query Attention Guided Inference}
\label{sec3.5}
Due to the lack of $st$ and $ed$ annotations, \textit{weakly supervised VMR} methods often compromise to designing two independent pipelines for training and inference. Under \textit{weakly supervised VMR}, the most common inference strategy is to select the best proposal from a series of proposals generated by methods like sliding window. Despite that it is still not possible to design a unified pipeline that handles training and inference  consistently under glance annotation, we propose to use a \textit{Query Attention Guided Inference} (QAGI) to best align the inference process to our aforementioned training strategy.

As illustrated in Figure \ref{fig:eval}, given a video $V$ and query $Q$, we first extract the features $\mathbf{v} \in \mathbb{R}^{L_v \times d_\text{model}}$ and $\mathbf{s} \in \mathbb{R}^{d_\text{model}}$ via the trained cross-modal representation module as described in previous sections. After that, we select an anchor point index $ap\in \{1,2...,L_v\}$ with the guidance of the query to video attention distribution. Specifically, the frame where the attention value reaches its maximum is chosen as the anchor frame, given by

\begin{align}
    ap=\operatorname*{arg\,max}_i\, \mathbf{a}_i.
\end{align}
A series of proposals are then generated around $ap$, \textit{i.e.}, we first apply a naive sliding window on the $L_v$ frames to generate a proposals pool $\{p_{i:j}\}$, then prune out all proposals that does not satisfy $i\leq ap \leq j$. On this basis, the proposal that maximizes the similarity score to the query is select as our final output, given by

\begin{align}
    \operatorname*{arg\,max}_{i,j}\, \text{max\_pool}([\mathbf{v}_{i:j}])^\top \mathbf{s}.
\end{align}

\section{Experiments}
To validate our proposed glance annotation and the method ViGA, extensive experiments are conducted on three publicly available datasets. We also perform ablation studies on different components in ViGA to investigate their influence in details. 

\subsection{Datasets}
We re-annotate the following datasets to fit in our proposed glance annotation. Specifically, we augment each example with a uniformly sampled timestamp $g$ in range $[st,ed]$.

\paragraph{\textbf{ActivityNet Captions}}
Krishna \textit{et al.} \cite{krishna2017dense} annotated the ActivityNet v1.3 dataset \cite{caba2015activitynet} which was originally designed for video captioning, and released the ActivityNet Captions dataset for VMR. It contains 19,994 YouTube videos from diverse domains. Following previous studies \cite{zhang2019cross,zhang2020learning}, we use the pre-defined split \textit{val\_1} as validation set and test on \textit{val\_2}. As a result, 37,421, 17,505, and 17,031 annotations are used for training, validating, and testing, respectively.

\paragraph{\textbf{Charades-STA}}
Gao \textit{et al.} \cite{gao2017tall} annotated the Charades dataset \cite{sigurdsson2016hollywood} using a semi-automatic approach and formed the Charades-STA dataset. It contains 9,848 videos of daily indoors activities. We follow the standard split of 12,408 and 3,720 annotations for training and testing defined by the annotator.

\paragraph{\textbf{TACoS}}
Regneri \textit{et al.} \cite{tacos2013} annotated the MPII Cooking Composite Activities dataset \cite{rohrbach2012script} which was originally designed for activity recognition, and formed the TACoS dataset. It contains 127 videos of cooking. We follow the standard split provided by \cite{gao2017tall}, and 9790, 4436, and 4001 annotations are included in training, validation and test set, respectively.
\begin{table}[t]
    \centering
    \begin{tabular}{c|ccc|c}
\hline
\multirow{2}{*}{Variants} & \multicolumn{3}{c|}{R@IoU=} & \multirow{2}{*}{mIoU}\\
                          & 0.3       & 0.5       & 0.7  &    \\ \hline
Video-NCE               & 35.58        & 18.30         & 8.54    &  25.34  \\
Clip-NCE & 16.72 & 6.25 & 2.02 & 14.93 \\
GLS-NCE                & 59.61   &   35.79    &   16.96    &    40.12   \\ \hline
\end{tabular}
    \caption{Ablation comparison among training with different NCE loss functions on ActivityNet Captions dataset.}
    \label{tab:GLS-NCE}
\end{table}

\begin{table}[t]
    \centering
    \begin{tabular}{c|ccc|c}
\hline
\multirow{2}{*}{Variants} & \multicolumn{3}{c|}{R@IoU=} & \multirow{2}{*}{mIoU}\\
                          & 0.3       & 0.5       & 0.7  &    \\ \hline
w/o QAG-KL               & 54.74         & 34.26        & 16.68    & 37.96   \\
w/ QAG-KL                & 59.61   &   35.79    &   16.96    &   40.12   \\ \hline
\end{tabular}
    \caption{Ablation comparison between model trained with and without QAG-KL loss on ActivityNet Captions dataset.}
    \label{tab:QAG-KL}
\end{table}

\begin{table}[t]
    \centering
    \begin{adjustbox}{width=0.45\textwidth}
    \begin{tabular}{c|ccc|c}
\hline
\multirow{2}{*}{Variants} & \multicolumn{3}{c|}{R@IoU=} & \multirow{2}{*}{mIoU}\\
                          & 0.3       & 0.5       & 0.7   &    \\ \hline
Sliding Window               & 58.13         & 31.23         & 13.62   &   38.24  \\
Query Attention Guided Inference & 59.61   &   35.79    &   16.96    &    40.12    \\ \hline
\end{tabular}
\end{adjustbox}
    \caption{Ablation comparison of inference with query attention guided inference / naive sliding window on ActivityNet Captions dataset.}
    \label{tab:eval_ablation}
\end{table}

\subsection{Evaluation Metric}
We evaluate our method using 1) recall of threshold bounded temporal intersection over union (R@IoU), which measures the percentage of correctly retrieved predictions where only the temporal IoU between the prediction and the ground truth greater than a certain threshold is accepted, and 2) mean averaged IoU (mIoU) over all predictions.

\subsection{Implementation Details}
We fix the 3D CNN modules for extracting visual features for a fair comparison. For all the three datasets, we use C3D as feature extractor. Since Charades-STA lacks a unified standard of feature extractor in previous studies, additional experiments using I3D and VGG features are also conducted for the completeness of comparison. For the word embedding, we adopt 840B GloVe for building a most complete vocabulary. To increase the capacity of our encoders, we stack two layers of our query, video and cross-modal encoders. The model dimension $d_\text{model}$ is set to 512, and the number of attention heads $h$ is set to 8 globally. Our model is trained with AdamW \cite{loshchilov2017decoupled} with a learning rate of 0.0001 half decaying on plateau. We clip the gradient norm to 1.0 during training. The batch size and $\sigma$ factor of the three datasets are empirically set to (256, 0.4), (256, 0.3) and (128, 1.0), respectively. All experiments are conducted on a Nvidia Tesla V100 GPU with 32GB memory.

\subsection{Ablation Studies} 

To evaluate the effectiveness of different components in our proposed ViGA, we conduct extensive ablation experiments on the ActivityNet Captions dataset.

\paragraph{\textbf{Effects of GLS-NCE}}
In order to verify the effectiveness of our proposed GLS-NCE loss (Equation \ref{eq:GLS-NCE}), we compare it with the aforementioned variants Video-NCE loss (Equation \ref{eq:Video-NCE}) and Clip-NCE loss (Equation \ref{eq:Clip-NCE}). The Video-NCE treats the video as a whole and maximizes the similarity between it and the text query. The Clip-NCE cuts a video into many clips, which increases the number of examples in the batch. However, as the distance between the clip and glance increases, its relevance to the query becomes lower. Therefore, our GLS-NCE assigns different weights to different clips according to their temporal distances to the glance. The results are listed in Table \ref{tab:GLS-NCE}. The performance of GLS-NCE is significantly ahead of others, thus showing its effectiveness. Besides, it is worthwhile to note that scores of Clip-NCE are almost half of Video-NCE, indicating that simply increasing the number of samples through clip segmentation is not beneficial, but sharply harms the performance instead. Comparing the three groups of experiments, we conclude that the performance improvement of GLS-NCE is not brought by increasing the number of examples by slicing a video into clips, while the enhancement from Gaussian label smoothing makes the main contribution.

\begin{figure}
    \centering
    \includegraphics[width=\linewidth]{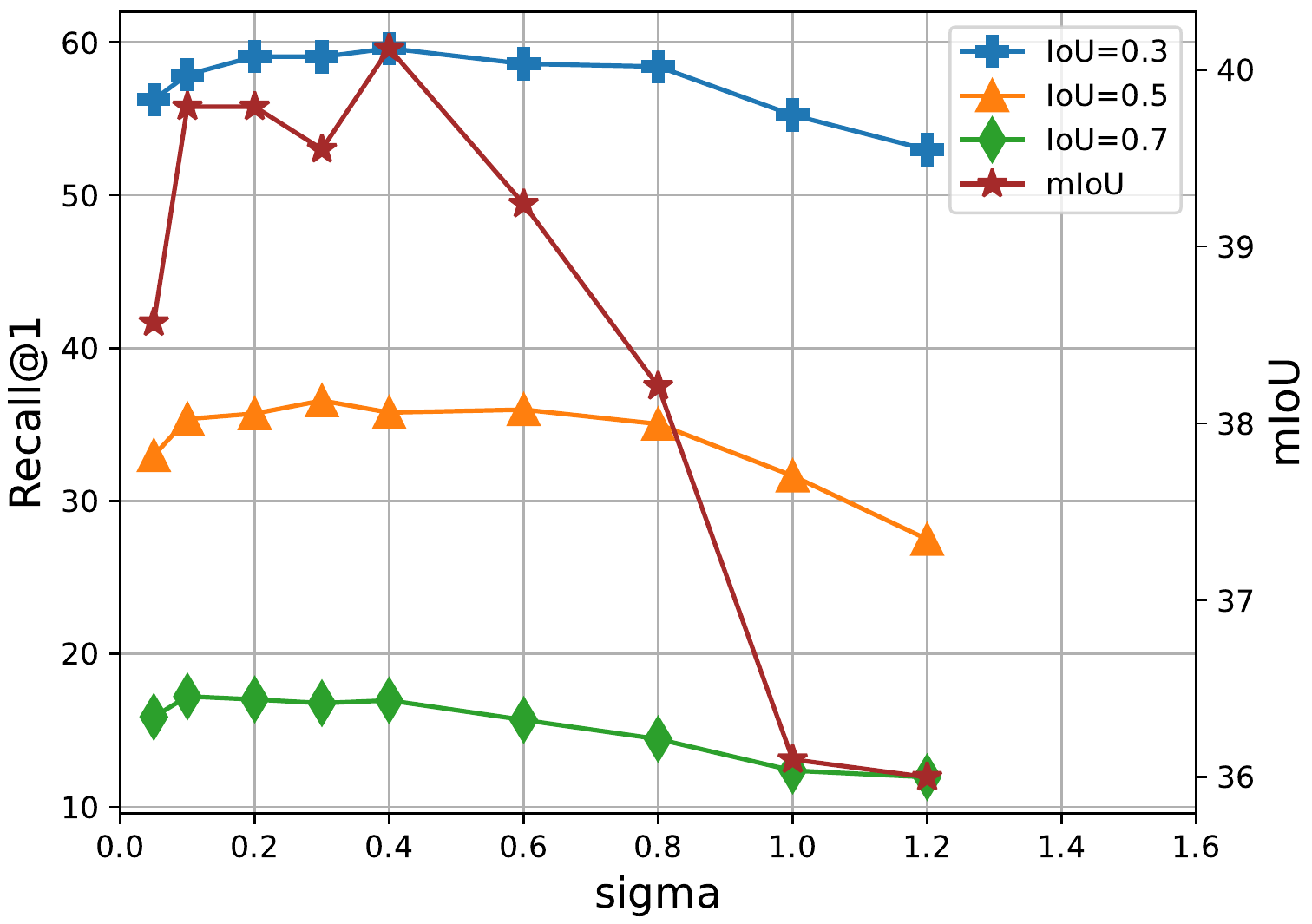}
    \caption{Influence of the hyperparameter $\sigma$ of the Gaussian distribution in Equation \ref{eq:gaussian} on ActivityNet Captions dataset.}
    \label{fig:sigma}
\end{figure}

\paragraph{\textbf{Effects of QAG-KL}}
The QAG-KL loss (Equation \ref{eq:QAG-KL}) encourages the model to pay more attention to the glance frame and its near neighbors in the training stage. To validate the its effectiveness, we conduct the ablation study of simply removing the QAG-KL loss. From the results in Table \ref{tab:QAG-KL}, we have the following observations. First, QAG-KL improves the moment retrieval performance on all evaluation metrics. This shows that in the training stage, QAG-KL can indeed make use of the prior information of glance annotation and help the model with better cross-modal alignment learning. Second, the performance with QAG-KL increases more significantly when the IoU threshold is 0.3 than other thresholds, reaching around $5\%$. We consider this gap is due to the fact that glance is a relatively weak prior information, so it performs better when the requirement of retrieval precision (reflected by the IoU) is not strict.

\paragraph{\textbf{Sliding Window \textit{vs.} Query Attention Guided Inference}}
To verify the effectiveness of our proposed QAGI, we evaluate the same trained model under different testing strategy, \textit{i.e.}, naive sliding window \textit{vs.} QAGI. The results in Table \ref{tab:eval_ablation} show that QAGI has advantages over the traditional sliding window based evaluation on all metrics. QAGI uses the attention matrix learned in the training stage to obtain the anchor frame for generating proposals in the test stage, which can filter out irrelevant proposals to a great extent, especially those with short durations. It is worthwhile to note that the improvement is more obvious under the metric with larger IoU threshold, as the performance raises by $4.5\%$ and $3.3\%$ respectively at IoU threshold of 0.5 and 0.7. This suggests that using the anchor is beneficial especially when the retrieval precision requirement is relatively high.

\paragraph{\textbf{Effects of the Gaussian distribution parameter $\boldsymbol{\sigma}$}}
In this ablation study, we focus on the hyperparameter $\sigma$ in Equation \ref{eq:gaussian}. Theoretically, $\sigma$ describes the dispersion degree of a Gaussian distribution: the larger the $\sigma$, the flatter the curve. In the context of our \textit{Gaussian Alignment Module}, the value of $\sigma$ controls to what extent that the weight at the glance frame which is always 1.0 disperses to other frames, hence affecting the overall positiveness of all the clips in the video. Consider an extreme example, when $\sigma$ takes a very large value, all frames in the video are assigned with weights close to 1.0. This means that we take all clips almost equally positive, which reduces the learning to be approximately equivalent to the video-level MIL under weak supervision. Therefore, choosing an appropriate $\sigma$ is important. As reported in Figure \ref{fig:sigma}, as $\sigma$ increases, the performance of the four metrics first increases and then decreases. Specifically, when $\sigma$ is set to 1.2, \textit{i.e.}, we over-assign positiveness to the clips, the performance of the four metrics decreases sharply (\textit{e.g.}, mIoU decreases from 40 to 36). On the other hand, when $\sigma$ is very small, \textit{i.e.}, we only take a very narrow range of video clips as important positive examples, the performance decreases because of losing some clips that are in fact informative positive examples (\textit{e.g.}, when $\sigma$ is set to 0.05, mIoU decreases by $2\%$). On the ActivityNet Captions dataset, the performance achieves its best when $\sigma$ is set to a medium value 0.4. This observation coincides with our theoretical analysis.

\subsection{Comparison with State of the Art}

\begin{table*}[ht]
\centering
\begin{adjustbox}{width=\textwidth}
\begin{tabular}{cc|cccc|cccc|cccc}
\hline
\multirow{2}{*}{Supervision} &
  \multirow{2}{*}{Method} &
  \multicolumn{4}{c|}{Charades-STA} &
  \multicolumn{4}{c|}{ActivityNet Captions} &
  \multicolumn{4}{c}{TACoS} \\ \cline{3-14} 
                                                       &            & R@0.3 & R@0.5 & R@0.7 & mIoU & R@0.3 & R@0.5 & R@0.7 & mIoU & R@0.3 & R@0.5 & R@0.7 & mIoU \\ \hline
\multicolumn{1}{c}{\multirow{5}{*}{\makecell[c]{Full \\ Supervision}}} & \multicolumn{1}{l|}{CTRL \cite{gao2017tall}} & -      & 23.63 & 8.89  & - & -     & -     & -   & - & 18.32 & 13.3  & -  & -   \\
\multicolumn{1}{c}{}                                  & \multicolumn{1}{l|}{QSPN \cite{xu2019multilevel}}      & 54.7  & 35.6  & 15.8  & - & 45.3  & 27.7  & 13.6 & - & -     & -     & - & -   \\
\multicolumn{1}{c}{}                                  & \multicolumn{1}{l|}{2D-TAN \cite{zhang2020learning}}    & -     & 39.7  & 23.31 & - & 59.45 & 44.51 & 26.54 & - & 37.29 & 25.32 & - & -    \\
\multicolumn{1}{c}{}                                  & \multicolumn{1}{l|}{LGI \cite{mun2020local}}       & 72.96 & 59.46 & 35.48 & 51.38 & 58.52 & 41.51 & 23.07 & 41.13 & -     & -     & -  & -    \\
\multicolumn{1}{c}{}                                  & \multicolumn{1}{l|}{VSLNet \cite{zhang2020span}}    & 70.46 & 54.19 & 35.22 & 50.02 & 63.16 & 43.22 & 26.16 & 43.19 & 29.61 & 24.27 & 20.03 & 24.11 \\ \hline
\multicolumn{1}{c}{\multirow{6}{*}{\makecell[c]{Weak \\ Supervision}}} & \multicolumn{1}{l|}{TGA \cite{mithun2019weakly}}       & 32.14 & 19.94 & 8.84  & - & -     & -     & -   & -  & -     & -     & -  & -   \\
\multicolumn{1}{c}{}                                  & \multicolumn{1}{l|}{SCN \cite{lin2020weakly}}       & 42.96 & 23.58 & 9.97 & - & 47.23 & 29.22 & -  & -   & -     & -     & - & -    \\
\multicolumn{1}{c}{}                                  & \multicolumn{1}{l|}{BAR \cite{wu2020reinforcement}}       & 44.97 & 27.04 & 12.23 & - & 49.03 & 30.73 & -  & -   & -     & -     & -  & -   \\
\multicolumn{1}{c}{}                                  & \multicolumn{1}{l|}{VLANet \cite{ma2020vlanet}}    & 45.24 & 31.83 & 14.17 & - & -     & -     & - & -    & -     & -     & - & -    \\
\multicolumn{1}{c}{}                                  & \multicolumn{1}{l|}{MARN \cite{song2020weakly}}      & 48.55 & 31.94 & 14.81 & - & 47.01 & 29.95 & -  & -   & -     & -     & -   & -  \\
\multicolumn{1}{c}{}                                  & \multicolumn{1}{l|}{LoGAN \cite{tan2021logan}}     & 51.67 & 34.68 & 14.54 & - & -     & -     & -  & -   & -     & -     & -  & -   \\
\multicolumn{1}{c}{}                                  & \multicolumn{1}{l|}{CRM \cite{huang2021cross}}       & 53.66 & 34.76 & 16.37 & - & 55.26 & 32.19 & -   & -  & -     & -     & - & -    \\ \hline
\multicolumn{1}{c}{\multirow{5}{*}{\makecell[c]{Glance \\ Supervision}}} & \multicolumn{1}{l|}{2D-TAN \cite{zhang2020learning} \dag} & - & - & - & - & 11.26 & 5.28 & 2.34 & - & 13.97    & 5.50     & 1.60 & - \\
\multicolumn{1}{c}{}                                  & \multicolumn{1}{l|}{LGI \cite{mun2020local} \dag}      & 51.94 & 25.67 & 7.98  & 30.83 & 9.34 & 4.11  & 1.31  & 7.82 & -    & -     & - & -    \\
\multicolumn{1}{c}{}                                  & \multicolumn{1}{l|}{ViGA (C3D)}  &   56.85    &   35.11    &    15.11   &   36.35    &   \textbf{ 59.61}   &   \textbf{35.79}    &   \textbf{16.96}    &    \textbf{40.12}   &  \textbf{19.62}    & \textbf{8.85}&  \textbf{3.22}& \textbf{15.47} \\
\multicolumn{1}{c}{}                                  & \multicolumn{1}{l|}{ViGA (VGG)} & 60.22 & 36.72 & 17.20 & 38.62 & - & - & - & - & - & - & - & - \\
\multicolumn{1}{c}{}                                  & \multicolumn{1}{l|}{ViGA (I3D)} & \textbf{71.21} & \textbf{45.05} & \textbf{20.27} & \textbf{44.57} & - & - & - & - & - & - & - & - \\ \hline
\end{tabular}
\end{adjustbox}
\caption{Performance comparison with the state-of-the-art methods under different supervision settings. ``\dag'' denotes our re-implemented results of fully supervised methods under glance annotations. In order to align with their original design, we give a relaxed glance condition by shrinking the original annotations to a random 3-seconds duration instead of one instant timestamp as in our results.}
\label{tab:sota}
\end{table*}

We compare the proposed ViGA with both fully and weakly supervised methods, which are introduced as follows. 

\paragraph{\textbf{Compared Methods}}
As shown in Table \ref{tab:sota}, we divide the compared methods into three sections according to the supervision types, including full supervision, weak supervision and glance supervision. When selecting methods from the literature to compare, we follow the rule of diversely selecting representative methods in different categories as introduced in Section \ref{sec:2} for the completeness of the comparison. For two-stage fully supervised methods, CTRL \cite{gao2017tall} is sliding window based and QSPN \cite{xu2019multilevel} is proposal based. In the end-to-end fully supervised methods, 2D-TAN \cite{zhang2020learning} belongs to anchor based, while LGI \cite{mun2020local} and VSLNet \cite{zhang2020span} are anchor free. 
For weak supervision, a dominant number of methods adopt MIL strategy than query reconstruction. Therefore, we select MIL-based methods like TGA \cite{mithun2019weakly}, VLANet \cite{ma2020vlanet}, LoGAN \cite{tan2021logan}, CRM \cite{huang2021cross} and  one representative reconstruction-based method SCN \cite{lin2020weakly}. 
Currently, CRM is the state of the art in \textit{weakly supervised VMR}. 

In addition to these existing studies, we apply glance annotation to two well-recognized fully supervised methods (\textit{i.e.}, 2D-TAN and LGI) for a more direct comparison to our proposed ViGA. In order to align with their original design, we give a relaxed glance condition by shrinking the original annotations to a random 3-seconds duration instead of one instant timestamp. Practically, we achieve this by directly changing the annotations in the data and run their publicly available source codes.

\begin{figure*}
    \centering
    \includegraphics[width=\textwidth]{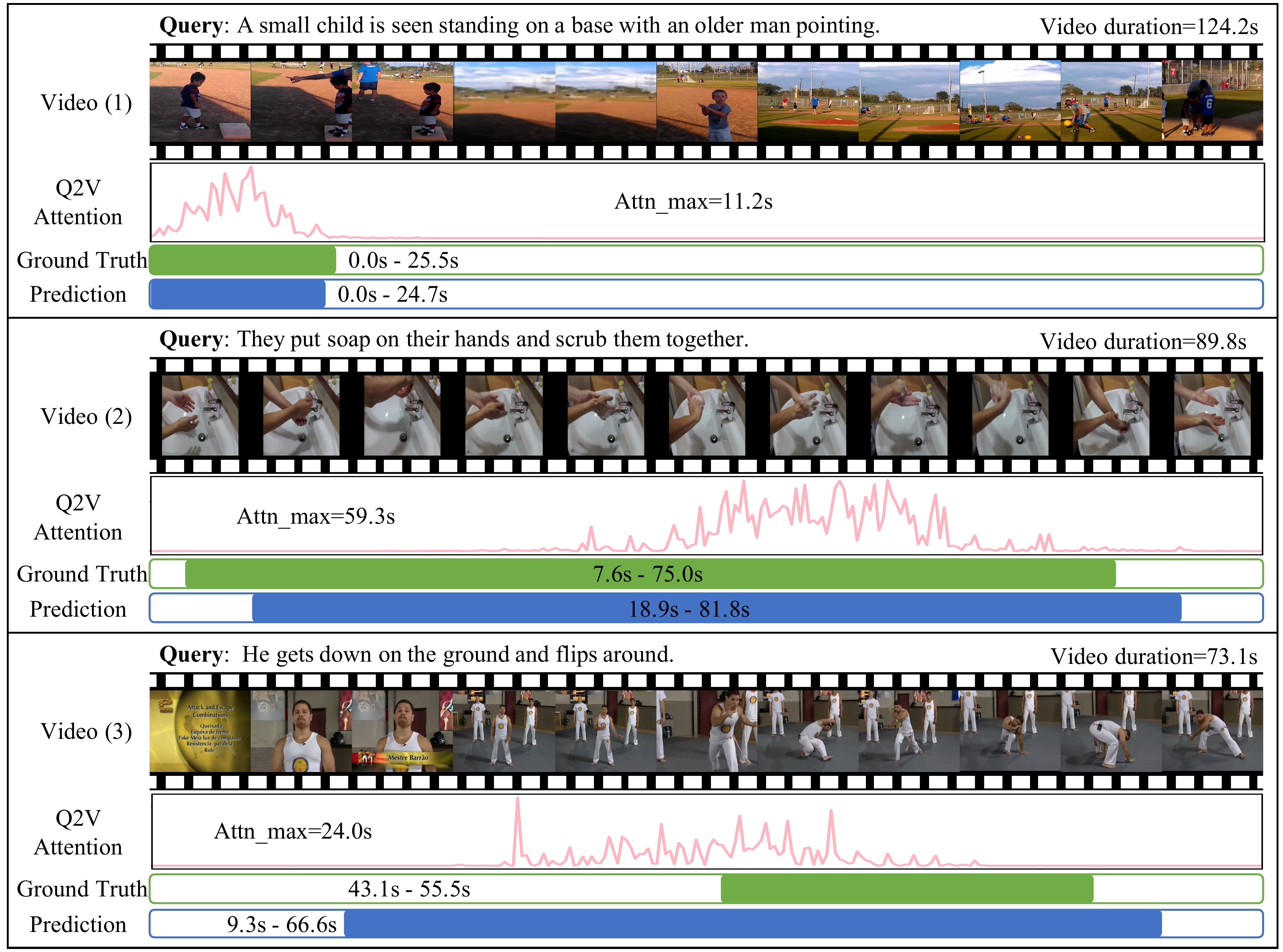}
    \caption{Some visualized examples from the test split of ActivityNet Captions. The first two examples are successful and the third is a failing case.}
    \label{fig:visualization}
\end{figure*}

\paragraph{\textbf{Observations and Discussions}} 
According to the results in Table \ref{tab:sota}, we can make a number of observations worthy discussing. 

1. In terms of all metrics, our proposed approach significantly exceeds the methods under weak supervision on the Charades-STA and ActivityNet Captions dataset. We improve the recall by $7\%$, $11\%$ and $4\%$ on Charades-STA when IoU is 0.3, 0.5 and 0.7, respectively. On ActivityNet Captions, the improvement is $5\%$ and $4\%$ when IoU is 0.3 and 0.5, respectively. We believe that on one hand, it shows that the setting of glance annotation is reasonable and has good potential in performance, and on the other hand, it also shows that ViGA succeeds in exploiting the information provided by glance annotation. In addition, in order to make ViGA standard and concise, we did not use some effective tricks in weak supervision methods, such as surrogate proposal selection in VLANet and temporal semantic consistency in CRM. This may take the performance of ViGA further, and we leave this as future work. 

2. When comparing to some fully supervised methods, we are surprised to find that when IoU is small (\textit{e.g.}, 0.3), our method almost reaches a same performance level. For example, on Charades-STA, our R@1 IoU=0.3 is $71.21\%$, $1.6\%$ lower than LGI and $0.8\%$ higher than VSLNet. On ActivityNet Captions, the recall is $59.61\%$, $1.9\%$ higher than LGI and $3.6\%$ lower than VSLNet. This suggests that under the scenario of coarse-grained retrieval requirements, glance annotation might be more advantageous than full annotation with acceptable performance yet significantly lower cost. However, there is still a lot of space for improvement when a high retrieval precision is required. For example, when the R@1 IoU=0.7, the performance gap between ViGA and LGI on Charades-STA reaches $15.21\%$. 

3. For the previously mentioned fully supervised method re-implemented under relaxed glance annotation, we have the following findings.  First, although we relax the setting of glance to 3 seconds, our approach shows superior performance in all three datasets. Second, we observe that the performance of re-implemented methods on ActivityNet Captions is not sufficiently satisfying. Therefore, it might be inadvisable to transplant fully supervised methods to glance annotation setting by directly changing the annotation to a instant moment or a short time duration, especially for the dataset like ActivityNet Captions, which generally has a long video duration and a wide range of moment lengths. 

4. As can be seen from Table \ref{tab:sota}, weakly supervised methods are often not tested on TACoS dataset because the videos in TACoS are very long and the moments to be retrieved are too short, \textit{i.e.}, the requirement of retrieval precision is very high. It might be hard for existing weakly supervised methods to deal with this situation. Our proposed ViGA shows positive in such case with a similar result to early fully supervised methods, such as CTRL.

\subsection{Qualitative Analysis}

Figure \ref{fig:visualization} shows some qualitative examples from the test split of ActivityNet Captions dataset, in which the green bar is the ground truth temporal boundary of the language query and the blue bar represents the predicted boundary of ViGA. We also visualize the query-to-video attention (pink curve under the video flow) to illustrate our proposed QAG-KL loss and query attention guided inference. Video (1) and Video (2) are successfully retrieved samples with high IoU. They show the effectiveness of our method from two aspects. For video (1), the video duration is very long (up to 124.2 seconds) and the moment to be retrieved is relatively short (25.5 seconds), which reveals that our proposed approach based on glance annotation can locate precisely when the video semantics is complex. As can be seen from the plot, this is benefited from a reasonable query-to-video attention distribution which is precisely positioned in the correct moment interval. On one hand, it enhances the cross-modal representation learning, and on the other hand, it provides a good anchor frame for inference. For video (2), we observe that ViGA successfully retrieves this long moment of nearly one minute. Given that we might be able to have good results of retrieving long segments under single frame glance annotation, it is reasonable to conjecture that the length of full annotation could have been reduced, even not to the extreme of just one single frame. Therefore, our qualitative results are in favor of the great potential of glance annotation. Inevitably, there are also failing cases. For example, in Video (3), the language query corresponds to a short clip of the man gets down on the ground and flips around, but our model recalls a long range segment containing the man, including a large part of the man standing, showing a lack of sufficient understanding of the fine-grained textual semantics. We consider that this is the hard part in the task of retrieving video moments with free-form text query. There is not sufficiently large amount of training data for learning fine-grained semantics because the free-form text query has great diversity. The model can be easily confused about some fine-grained actions, such as ``get down on the ground and flip around'' here.

\section{Limitations}
Due to our limited resource, we are only able to re-annotate the datasets in an automatic way by doing random sample in the time interval of original annotation instead of manually annotating them. Although we achieve good results in the previous experiments, there are some inevitable problems in this simple re-annotation strategy. For example, some queries might contain multiple semantics, which are not possible to be captured by only one glance. Also, in some rare cases, meaningless frames that would pollute the training data such as camera scene changes might be sampled as the glance, which could have been filtered out manually. We hope a manually annotated dataset in glance annotation could be collected in the future to support follow-up research in this direction.

\section{Conclusion}
In this paper, we study the problem of VMR. After analysing the advantages and limitations of the two existing VMR paradigms \textit{fully supervised VMR} and \textit{weakly supervised VMR}, we find that \textit{weakly supervised VMR} can be augmented with trivial cost, and propose a new data annotation paradigm named as glance annotation. Under glance annotation, we propose ViGA, a novel clip-level contrastive learning framework, as a pioneer method to solve this problem. Extensive experiments are conducted on three publicly available datasets ActivityNet Captions, Charades-STA and TACoS, and ViGA outperforms existing weakly supervised methods by a large margin. Therefore, we conclude that glance annotation is a promising new data annotation paradigm for VMR, and ViGA is a feasible method for glance annotated VMR. Our results support further research and applications of glance annotation in real-life problems.

\section*{Acknowledgement}
This research is conducted within the first two authors' internship in bilibili. We are grateful to bilibili AI for the support and inspiration.

\bibliographystyle{ACM-Reference-Format}
\bibliography{main}

\end{document}